\documentclass[runningheads]{llncs}
\usepackage{graphicx}
\usepackage{amsmath}
\usepackage{amssymb}
\usepackage{booktabs}
\usepackage{times}
\usepackage{epsfig}
\usepackage{flushend}
\usepackage{comment}
\usepackage{subcaption}
\usepackage{amsfonts,algorithm,algpseudocode}
\newcommand{\cmmnt}[1]{}

\newcommand{\etal}{\emph{et al. }}

\begin{document}
\title{Adaptive Weighted Co-Learning for\\ Cross-Domain Few-Shot Learning}

\author{Abdullah Alchihabi, Marzi Heidari,  Yuhong Guo}
\institute{Carleton University, Ottawa, Canada}

\maketitle              
%
\begin{abstract}
Due to the availability of only a few labeled instances for the novel target prediction task and the significant domain shift between the well annotated source domain and the target domain, cross-domain few-shot learning (CDFSL) induces a very challenging adaptation problem. In this paper, we propose a simple Adaptive Weighted Co-Learning (AWCoL) method to address the CDFSL challenge by adapting two independently trained source prototypical classification models to the target task in a weighted co-learning manner. The proposed method deploys a weighted moving average prediction strategy to generate probabilistic predictions from each model, and then conducts adaptive co-learning by jointly fine-tuning the two models in an alternating manner based on the pseudo-labels and instance weights produced from the predictions. Moreover, a negative pseudo-labeling regularizer is further deployed to improve the fine-tuning process by penalizing false predictions. Comprehensive experiments are conducted on multiple benchmark datasets and the empirical results demonstrate that the proposed method produces state-of-the-art CDFSL performance.

\end{abstract}

\section{Introduction}

Deep learning has achieved great success on a wide set of computer vision tasks, ranging from image classification to image segmentation and object detection. Such success however has been contingent on the availability of a large set of annotated training instances, which induces significant data annotation cost. 
To alleviate this annotation burden, Few-Shot Learning (FSL) methods have been developed to exploit a base set of classes with a large number of labeled instances to help train deep prediction models for target tasks over novel classes, where only a few instances are labeled for each class. 
Standard FSL methods are designed for an in-domain setting, where data for the base classes and the novel classes are from the same domain, and their performance degrades as the dissimilarity between the base dataset (source domain) and the novel target task (target domain) increases \cite{guo2020broader}, which raises significant demands for cross-domain few-shot learning.

Cross-Domain Few-Shot Learning (CDFSL) expands the FSL study by leveraging labeled base datasets from a remote source domain to help FSL in an annotation-strenuous target domain. CDFSL is much more challenging than its in-domain counterpart due to the limited few-shot labeled instances for the novel target prediction task and the significant domain shift between the source and target domains, and has just started gaining attention from the research community. 
Some approaches have been developed to tackle CDFSL by employing data augmentation and data generation techniques to increase the number of available labeled instances~\cite{advTaskAug,islam2021dynamic}, 
while several others have deployed ensemble learning or hierarchical variational memory techniques to learn diverse features and facilitate model adaptation from the source domain to the target domain~\cite{adler2020cross,du2022hierarchical}. 
However, these methods either have limited scalability for higher-shot learning problems~\cite{advTaskAug}, induce significant computational cost~\cite{adler2020cross,du2022hierarchical} or require a large set of unlabeled target-domain samples to be available during source-domain training \cite{islam2021dynamic}.

In this paper, we propose a simple but novel Adaptive Weighted Co-Learning (AWCoL) method for CDFSL. The proposed method jointly fine-tunes two simple prototypical classification models 
independently pre-trained on the source-domain dataset for the target few-shot prediction task in a weighted adaptive co-learning manner.
Specifically, for each model we propose to use a weighted moving average (WMA) strategy to generate robust probabilistic predictions for the query instances of the target task, where the class prototypes are computed using the support instances. Pseudo-labels and the corresponding confidence weights for the query instances can be determined based
on the average probabilistic predictions produced by the two models.
The two prototypical models are then fine-tuned in an alternating fashion 
on the pseudo-labeled query instances by minimizing a weighted cross-entropy loss, aiming to effectively exploit the more diverse unlabeled queries for stable co-learning. 
Moreover, we also adopt a negative pseudo-label regularizer to further assist the fine-tuning process by pushing the predictions away from the negative pseudo-labels. To evaluate the proposed method, we conduct comprehensive experiments on eight benchmark datasets for CDFSL. The proposed AWCoL method demonstrates superior performance compared to the 
existing state-of-the-art CDFSL methods. 

\section{Related Works}

\subsection{Few-Shot Learning}
Standard in-domain few-shot learning (FSL) methods can be broadly categorized into 
the following four main groups: generative and augmentation-based methods, transfer learning methods, 
meta-learning methods, and metric learning methods. 
The generative and augmentation-based methods generate new instances to increase the training set diversity
\cite{zhang2018metagan,lim2019fast}. 
The transfer learning methods focus on adapting models pre-trained on the base dataset
to the novel target task with fine-tuning techniques~\cite{jeong2020ood,guo2019spottune}. 
Meta-learning approaches learn to perform few-shot learning by simulating few-shot tasks on the annotated base dataset. For example,
MAML aims to learn good parameter initialization on the base classes in order to easily adapt the model to the target task~\cite{finn2017model};  
MetaOpt employs meta-learning to train linear classifier as base learners~\cite{lee2019meta}.  
Metric learning methods aim to exploit the base dataset to induce good similarity/distance metrics.  
In particular, MatchingNet~\cite{vinyals2016matching} uses attention and memory to enable rapid metric learning. 
ProtoNet~\cite{snell2017prototypical} represents each class using a prototype and learns a metric space in which instances are classified based on their distances to the class prototypes. 
Other metric learning approaches, such as RelationNet \cite{sung2018learning}, GNN \cite{garcia2018fewshot} and Transductive Propagation Network (TPN) \cite{liu2018learning}, utilize similarities between the support instances and query instances to perform few-shot learning.

Recently, a few works have exploited co-learning to enhance FSL \cite{xu2022co,yin2021hierarchical,chen2020diversity}. Xu \etal employs co-learning to train two simple classifiers to generate better pseudo-labels for the unlabeled samples and alleviate the labeled data scarcity limitation in FSL. Yin \etal proposed Hierarchical Graph Attention network (HGAT) to address the multi-modal FSL task, where attention-based co-learning is employed to model the inter-modality relationships \cite{yin2021hierarchical}. Diversity Transfer Network (DTN) \cite{chen2020diversity} generates diverse instances for novel class categories by composing the instances from the known categories, and then uses an auxiliary task co-training to stabilize the meta-training process.

\subsection{Cross-Domain Few-Shot Learning}

By broadening the applicable domains, cross-domain few-shot learning (CDFSL)
has recently started drawing more attention~\cite{guo2020broader}. 
Tseng \etal proposed a feature-wise transformation (FWT) layer, 
which employs affine transformations to augment the learned features, 
to help FSL models generalize across different domains~\cite{Tseng2020CrossDomain}. 
Li \etal proposed a ranking distance calibration with fine-tuning (RDC-FT) method 
that constructs a non-linear subspace to reduce task-irrelevant features \cite{li2022ranking}. 
The Hierarchical Variational neural Memory framework (HVM) 
developed in~\cite{du2022hierarchical} 
learns hierarchical prototypes with variational inferences.

Data augmentation and data generation techniques 
have also been used for CDFSL~\cite{advTaskAug,islam2021dynamic}. 
Adversarial task augmentation (ATA) has been deployed to 
improve meta-learning models for cross-domain few-shot classification \cite{advTaskAug}. 
Islam \etal proposes a dynamic distillation method 
for CDFSL with unlabeled data, which 
exploits data augmentations to match predictions between 
a student network and a teacher network on unlabeled data \cite{islam2021dynamic}. 
In addition, self-supervised learning based methods 
have also been developed for CDFSL
\cite{das2022confess}. 
Das \etal proposed a Contrastive learning and feature selection system (ConFeSS) to bridge the domain shift with self-supervised representation learning \cite{das2022confess}.
%

\section{The Proposed Method}

\begin{figure*}[t]
  \centering
 \includegraphics[width= 1 \textwidth]{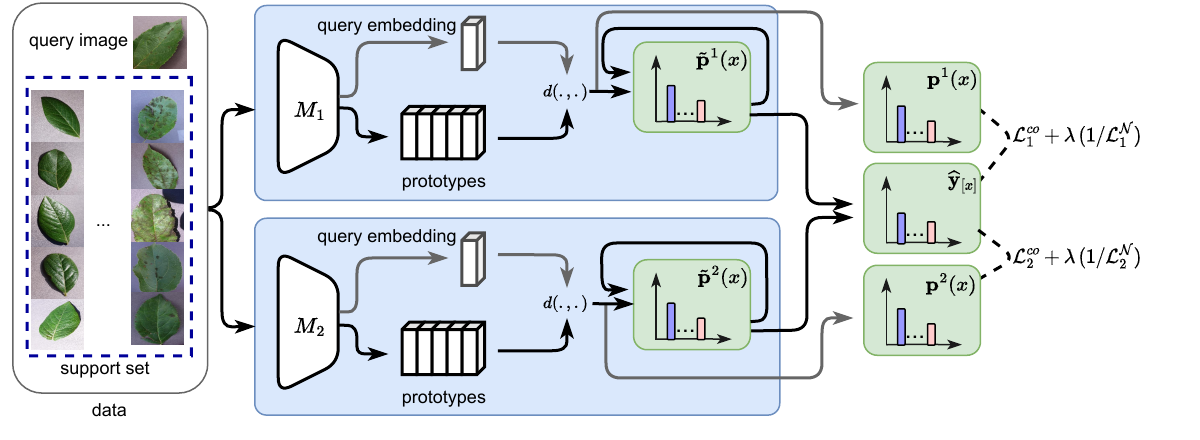} 
	 \caption{
The framework of the proposed Adaptive Weighted Co-Learning (AWCoL) method 
with two prototypical models, $M_1$ and $M_2$, pre-trained independently in the source domain: 
At each fine-tuning iteration in the target domain, 
each model generates probabilistic predictions for the query instances 
using a weighted moving average strategy. 
The predictions from both models are then combined to determine the positive pseudo-labels, negative pseudo-labels, and adaptive weights for the query instances, which are used to fine-tune the two models in the target domain in an adaptive co-learning manner. 
}
   \label{fig:target}
   \vskip -.10in
\end{figure*}

\paragraph{\bf Problem Setup}
CDFSL aims to train a model in an annotation-abundant source domain, 
and then adapt/fine-tune the model for the novel few-shot prediction task in the target domain. 
The source domain and the target domain are assumed to have different distributions in the input feature space ($\mathcal{P}_s \neq \mathcal{P}_t$) 
and disjoint classes ($\mathcal{Y}_s \cap \mathcal{Y}_t=\emptyset$) in the output label space. 
In the target domain, the model has access to a labelled support set $S=\{(x_i,y_i)\}^{N_s}_{i=1}$ and is evaluated on a query set $Q=\{(x_i,y_i)\}^{N_q}_{i=1}$, where $N_s$ and $N_q$ are the sizes of the support and query sets respectively. 
The support set typically consists of $K$ instances from each of the $N$ classes, resulting a total of $N_s=NK$ labeled support instances. This setup is commonly dubbed as N-way K-shot learning. 

In this section, we present the proposed Adaptive Weighted Co-Learning (AWCoL) method 
to address the CDFSL problem. 
The overall framework of AWCoL is illustrated in Figure \ref{fig:target}. 
The method provides a simple adaptive co-learning mechanism for fine-tuning
two models independently pre-trained in the source domain. 
In particular, 
we consider pre-training two simple prototypical FSL models in the source domain.
The models are then fine-tuned for the target FSL task in the target domain
using an adaptive weighted co-learning procedure. 
In each iteration of the fine-tuning,  
a weighted moving average (WMA) strategy is applied to each of the two models independently
to generate probabilistic predictions for the unlabeled query instances. 
The predictions from the two models are then combined to produce 
pseudo-labels and adaptive weights
for the query instances, which are subsequently used 
to fine-tune the models in an adaptive co-learning manner. 
In addition, negative pseudo-labels generated from the predictions
can also be used to enhance the fine-tuning process 
by pushing the models away from the predicted negative class labels.  
Overall, the two models are fine-tuned in an alternating 
fashion to yield a stable co-learning process,
which will be elaborated below.

\subsection{Source-domain Pre-training}
We use a simple prototypical FSL model~\cite{snell2017prototypical} as our base classification model,
due to its 
simplicity and efficiency. 
In the well-annotated source domain, we pre-train two classic prototypical few-shot learning models independently 
using two different sets of randomly sampled FSL tasks, 
which we refer to as $M_1$ and $M_2$. 
Each of the two models 
has its own feature encoder $f$ parameterized by $\theta$ 
that maps an input image $x$ to 
a feature embedding vector. 
For each model, given each sampled training FSL task 
with a support set $S$ and a query set $Q$, 
a prototype embedding vector 
$\mathbf{c}_n$ can be computed 
from the support instances for each class $n$ using the feature extractor $f$, 
such that: 
\begin{equation}\label{eq:calc_proto}
  \mathbf{c}_n = \frac{1}{K} \sum\nolimits_{(x,y) \in S_n}f_{\theta}(x),
\end{equation}
where $S_n$ denotes the set of $K$ support instances from class $n$. 
With the class prototypes, 
the prototypical classifier predicts the class probability vector
$\mathbf{p}(x)=[\mathbf{p}_1(x), \cdots, \mathbf{p}_N(x)]$
for each query instance $x$ by calculating
the negative distances between $f_\theta(x)$ and each class prototype vector
and then normalizing them using the softmax function,
such that:
\begin{equation}\label{eq:calc_dist}
   \mathbf{p}_j(x)
	=\frac{\exp(-d(f_\theta(x),\mathbf{c}_{j}))}{\sum_{n=1}^{N}\exp(-d(f_{\theta}(x),\mathbf{c}_n))},
\end{equation}
where $d(\cdot,\cdot)$ is a distance function and 
$\mathbf{p}_j(x)$
is the predicted probability that 
instance $x$ belongs to class $j$.

Each model can be trained consecutively on the set of randomly sampled FSL tasks in the source domain
by minimizing the following cross-entropy loss over the query instances:
\begin{equation}
	\mathcal{L}_{CE}(Q) =  \sum\nolimits_{(x,y)\in Q} \ell_{CE}(\mathbf{y}_y, \mathbf{p}(x)),
\end{equation}
where $\ell_{CE}$ is the cross-entropy loss function, 
$\mathbf{p}(x)$ and $\mathbf{y}_y$ 
are the predicted class probability vector and the ground-truth one-hot label indicator vector 
(with value 1 at the $y$-th entry) 
respectively for the query instance $x$. 
The output of the source-domain pre-training step is two 
independently trained prototypical classification models,
$M_1$ and $M_2$, with separate feature encoders $f_1$ and $f_2$ parameterized by $\theta_1$ and $\theta_2$ respectively.

\subsection{Weighted Moving Average based Probabilistic Prediction Generation}
Given the significant domain shift between the source domain and target domain in the CDFSL problem, 
it is important to adapt the models pre-trained in the source domain to the novel FSL task in the target domain
through fine-tuning. 
Due to the limited number of labeled support instances, $S$, available for the FSL target task, 
we propose to exploit the unlabeled query instances, $Q$, with pseudo-labels for transductive model fine-tuning, 
aiming to increase training data diversity and avoid over-adaptation.

The pseudo-labels for query instances will be produced based on the predictions
of the two models, $M_1$ and $M_2$.
In order to generate robust predictions across fine-tuning iterations, 
we adopt a weighted moving average (WMA) strategy to produce probabilistic predictions
for the query instances using each of the two models separately. 
Specifically, let $M_{m}$ denote either one of the two models, such that $m\in\{1,2\}$.
Then at each fine-tuning iteration, we calculate the prototype vector for each class 
from the support instances with the feature extractor $f_m$
in the given model $M_{m}$ using Eq.~(\ref{eq:calc_proto}). 
Next for each query instance $x$, 
we calculate its class prediction probability vector ${\bf p}^{m}(x)$
using Eq.~(\ref{eq:calc_dist})
under the current model $M_m$.
Instead of using this predicted probability vector directly, 
we maintain a WMA class prediction probability vector
$\tilde{\mathbf{p}}^m(x)$ 
for each query instance $x$ across the fine-tuning iterations,
which can be updated in each iteration using the new predictions ${\bf p}^{m}(x)$ as follows:
\begin{equation}
	\tilde{\mathbf{p}}^{m}(x)=   
	(1-\alpha_m) \;   \tilde{\mathbf{p}}^{m}(x)
	+\alpha_m \; \mathbf{p}^{m}(x) 
	\label{eq:M1_WMA}
\end{equation}
where $\alpha_m\in(0,1)$
is a trade-off hyper-parameter that controls the combination weights between 
the class prediction probability vector of the current iteration $\mathbf{p}^{m}$ 
and the WMA class prediction probability vector $\tilde{\mathbf{p}}^{m}$ from the previous iteration,
and hence determines the updating degree. 
With such
a WMA update, we expect to maintain stable improvements 
for pseudo-label predictions over the query instances and avoid local oscillations. 

To further improve the stability of the WMA prediction generation, 
we employ a rectified annealing schedule to 
update the hyper-parameter $\alpha_m$ in each iteration as follows:  
\begin{equation}
    \alpha_m = \max(\alpha_{\text{min}}, \gamma\,\alpha_m),
	\label{eq:alpha}
\end{equation}
where $\alpha_{\text{min}}$ 
provides a lower bound for $\alpha_m$,
and $\gamma\in(0,1)$ is a reduction ratio hyper-parameter 
that gradually reduces the $\alpha_m$ value across iterations.
This annealing schedule allows the method 
to perform larger updates to the WMA class prediction probability vectors 
with larger $\alpha_m$ values
in the early fine-tuning iterations, 
while decreasing
the updating degree with smaller $\alpha_m$
in later iterations of fine-tuning,
aiming to help the fine-tuning process reach convergence.

\subsection{Weighted Adaptive Co-Learning}

With the WMA class prediction probability vectors for the query instances produced by $M_1$ and $M_2$, 
we propose to fine-tune the two models together using a weighted adaptive co-learning procedure 
by effectively integrating their predictions. 
Specifically, we simply take a softmax rescaled average of the WMA predictions from the two models
to produce the co-prediction probability vector $\tilde{\mathbf{p}}^{\text{co}}(x)$  
for each query instance $x$, such that:
\begin{equation}\label{eq:AVG_WMA}
	\tilde{\mathbf{p}}^{\text{co}}(x) =    
	\text{softmax} (\text{avg}(\, \tilde{\mathbf{p}}^{1}(x)\, ,\,\tilde{\mathbf{p}}^{2}(x)\, )).
\end{equation}
Here we apply the softmax function to rescale the average probability vector 
towards less skewed predictions, 
aiming to avoid early stage poor local optima for the co-learning procedure 
and help stabilize the fine-tuning process.

The pseudo-labels for the query instances can then be determined based on the integrated
co-prediction probability vectors. 
Specifically, for a query instance $x$, its pseudo-label $\widehat{y}_{[x]}$ will be 
a scalar index indicating the class with the highest prediction probability: 
\begin{equation}\label{eq:K}
	\widehat{y}_{[x]}=\text{argmax}_{n\in\{1,\cdots,N\}}\; \tilde{\mathbf{p}}_n^{\text{co}}(x) 
\end{equation}
where $N$ is the total number of classes for the target FSL task. 
For convenience, we further transform the pseudo-label $\widehat{y}_{[x]}$
into a one-hot pseudo-label indicator vector $\widehat{\bf y}_{[x]}\in \{0,1\}^N$
that has a single 1 value at its $\widehat{y}_{[x]}-$th entry. 

By producing pseudo-labels from the integrated predictions of the two models,
we expect to yield more accurate pseudo-labels than using each individual model alone. 
Nevertheless, noisy labels, i.e., mistakes, are still unavoidable among the predicted pseudo-labels. 
To alleviate the negative impact of the potential pseudo-label noise,
we further propose to calculate a dynamic and adaptive weight 
for each unlabeled query instance $x$ based on the current co-predictions:
\begin{equation}\label{eq:W}
	w_{[x]} =  \max_{n\in\{1,\cdots,N\}}\; \tilde{\mathbf{p}}_n^{\text{co}}(x)    
\end{equation}   
This adaptive weighting mechanism allows us 
to assign different weights to different query instances 
based on the co-prediction confidence of the generated pseudo-labels,
such that the query instances with more confident pseudo-labels---less likely to be noise---get larger weights.

With the co-predicted pseudo-labels and adaptive weights for the query instances,
we fine-tune the two models using an alternating co-learning procedure; 
i.e., alternatively updating each model across co-learning iterations. 
In a given iteration, for the chosen model $M_m$ ($m\in\{1,2\}$), we update its model parameters $\theta_m$
by minimizing the following adaptive weighted cross-entropy loss on the pseudo-labeled query instances:
\begin{equation}\label{eq:M1_Q_pos}
	\mathcal{L}^{\text{co}}_{m}   = 
	\frac{1}{\sum_{x \in Q } w_{[x]}} \sum_{x \in Q}  w_{[x]}\, 
	\ell_{CE}\!\left(\widehat{\mathbf{y}}_{[x]}, \,\mathbf{p}^{m}(x)\right)  
\end{equation}
where $\mathbf{p}^{m}(x)$ is the prediction probability vector for instance $x$ using 
the prototypical model $M_m$ via Eq.~(\ref{eq:calc_dist}).

\subsection{Negative Pseudo-Label Regularization}
In addition to exploiting the predicted positive pseudo-labels, 
we further propose to employ the negative pseudo-labels---i.e., the other class labels
except the predicted ones---to regularize the co-learning process. 
Specifically, for a query instance $x$, in addition to determining
the predicted positive pseudo-label $\widehat{y}_{[x]}$ using Eq.~(\ref{eq:K})
from the integrated co-prediction probability vector $\tilde{\mathbf{p}}^{co}(x)$,
we treat all the other classes 
$\{1,\cdots, N\}/\widehat{y}_{[x]}$
as candidate negative labels for $x$ and randomly choose one 
to use as a negative pseudo-label: 
\begin{equation}\label{eq:K_bar}
	\widehat{y}^{\mathcal{N}}_{[x]} =  \text{Rand}(\{1,..,N\}/\ \widehat{y}_{[x]})
\end{equation}
Same as before, we can transform this negative pseudo-label 
index into a one-hot {negative pseudo-label indicator vector} 
$\widehat{\bf y}^{\mathcal{N}}_{[x]}\in \{0,1\}^N$
with a single 1 at its $\widehat{y}^{\mathcal{N}}_{[x]}-$th entry. 

For a multi-class classification problem, only one class label is the true label
for each given query instance. 
Our hypothesis is that
even when the predicted positive pseudo-label is a false positive label,
a randomly selected negative pseudo-label can still possibly be a true negative label. 
Hence by minimizing the likelihood of the chosen negative pseudo-labels on
query instances, one can push the model away from making true negative predictions
and help alleviate the negative impact of the prediction noise in the positive pseudo-labels. 
To encode this hypothesis, we incorporate the selected negative pseudo-labels into 
the co-learning process by {\em maximizing} the following adaptive weighted cross-entropy loss 
on the query instances for each model $M_m$ ($m\in\{1,2\}$):
\begin{equation}\label{eq:M1_Q_neg}
	\mathcal{L}^{\mathcal{N}}_{m}   = \frac{1}{\sum_{x \in Q } {w}_{[x]}} 
	\sum_{x \in Q}  w_{[x]}\, \ell_{CE}\!\left(\widehat{\mathbf{y}}^{\mathcal{N}}_{[x]},\, \mathbf{p}^{m}(x)\right)  
\end{equation}
where the adaptive weight $w_{[x]}$ and the label prediction probability vector $\mathbf{p}^{m}(x)$
are same as 
the ones used for positive pseudo-labels
in Eq.~(\ref{eq:M1_Q_pos}).

\subsection{Alternating Adaptive Co-Learning}

By integrating both the co-learning minimization loss in Eq.~(\ref{eq:M1_Q_pos})
and the negative pseudo-label based maximization loss in Eq.~(\ref{eq:M1_Q_neg})
together, we deploy the following adaptive co-learning loss
to fine-tune each model $M_m$ ($m\in\{1,2\}$):
\begin{equation}\label{eq:M1_all}
	\mathcal{L}_{m} =    \mathcal{L}^{co}_{m} + \lambda \, (1/ \mathcal{L}^{\mathcal{N}}_{m}) 
\end{equation}
where $\lambda$ is a trade-off hyper-parameter 
that controls the relative contribution of the two loss terms.
Minimizing this total loss 
will simultaneously minimize the positive pseudo-label based weighted adaptive co-learning loss $\mathcal{L}^{co}_{m}$
and maximize the negative pseudo-label based weighted cross-entropy loss $\mathcal{L}^{\mathcal{N}}_{m}$.

In order to maintain 
a stable fine-tuning process and prevent oscillating co-prediction updates 
between the two models,
we propose to utilize an alternating updating mechanism to fine-tune the two models
with the integrated total loss. 
In particular, we consecutively update each model for $\beta$ iterations while keeping the 
other model fixed, and then switch to update the other model similarly.
As a result, when calculating the co-prediction probability vector $\tilde{\mathbf{p}}^{\text{co}}(x)$
in each iteration, only one model's WMA predictions will be updated 
while anchoring on the fixed WMA predictions of the other model. 
%

\section{Experiments}

\subsection{Experimental Setup}

\begin{table*}[t]
\caption{Mean classification accuracy (95\% confidence interval in brackets) 
	on 8 target domain datasets for cross-domain 5-way 5-shot classification.  
	``$^\ast$'' and ``$^\dagger$'' indicate the results reported in \cite{guo2020broader} and \cite{advTaskAug} respectively.} 
\setlength{\tabcolsep}{2pt}	
\resizebox{\textwidth}{!}{

\begin{tabular}{l|l|l|l|l|l|l|l|l}
\hline	
 & ChestX                    & CropDisea.                & ISIC                       & EuroSAT                  & Places                   & Planatae                    & Cars                     & CUB                      \\
   \hline 
	MatchingNet$^\ast$\cite{vinyals2016matching} & $22.40_{(0.70)}$  & $66.39_{(0.78)}$ & $36.74_{(0.53)}$  & $64.45_{(0.63)}$     & \multicolumn{1}{|c|}{$-$}          &      \multicolumn{1}{|c|}{$-$}                         &     \multicolumn{1}{|c|}{$-$}                        &    \multicolumn{1}{|c}{$-$}                         \\
MAML$^\ast$\cite{finn2017model} & $23.48_{(0.96)}$    & $78.05_{(0.68)}$    & $40.13_{(0.58)}$   & $71.70_{(0.72)}$   & \multicolumn{1}{|c|}{$-$}          &      \multicolumn{1}{|c|}{$-$}                         &     \multicolumn{1}{|c|}{$-$}                        &    \multicolumn{1}{|c}{$-$}                         \\
   ProtoNet$^\ast$\cite{snell2017prototypical}     & $24.05_{(1.01)}$  & $79.72_{(0.67)}$  & $39.57_{(0.57)}$        & $73.29_{(0.71)}$        & $58.54_{(0.68)}$        &     $46.80_{(0.65)}$                      &     $41.74_{(0.72)}$                       &  $55.51_{(0.68)}$                    \\
   MetaOpt$^\ast$\cite{lee2019meta} & $22.53_{(0.91)}$         & $68.41_{(0.73)}$   & $36.28_{(0.50)}$   & $64.44_{(0.73)}$& \multicolumn{1}{|c|}{$-$}          &      \multicolumn{1}{|c|}{$-$}                         &     \multicolumn{1}{|c|}{$-$}                        &    \multicolumn{1}{|c}{$-$}                         \\

 RelationNet$^\dagger$\cite{sung2018learning}     & $24.07_{(0.20)}$          & $72.86_{(0.40)}$     & $38.60_{(0.30)}$ & $65.56_{(0.40)}$   & $64.25_{(0.40)}$          & $42.71_{(0.30)}$            & $40.46_{(0.40)}$     & $56.77_{(0.40)}$     \\

   
   GNN$^\dagger$\cite{garcia2018fewshot}      & $23.87_{(0.20)}$       & $83.12_{(0.40)}$  & $42.54_{(0.40)}$   & $78.69_{(0.40)}$        & $70.91_{(0.50)}$  & $48.51_{(0.40)}$ & $43.70_{(0.40)}$   & $62.87_{(0.50)}$ \\
   TPN $^\dagger$\cite{liu2018learning}  & $22.17_{(0.20)}$ & $81.91_{(0.50)}$ & $45.66_{(0.30)}$ & $77.22_{(0.40)}$ & $71.39_{(0.40)}$  & $50.96_{(0.40)}$  & $44.54_{(0.40)}$ & $63.52_{(0.40)}$ \\
   \hline
MatchingNet+FWT$^\ast$\cite{Tseng2020CrossDomain}  & $21.26_{(0.31)}$         & $62.74_{(0.90)}$           & $30.40_{(0.48)}$     & $56.04_{(0.65)}$         & \multicolumn{1}{|c|}{$-$}          &      \multicolumn{1}{|c|}{$-$}                         &     \multicolumn{1}{|c|}{$-$}                        &    \multicolumn{1}{|c}{$-$}                         \\
  ProtoNet+FWT$^\ast$\cite{Tseng2020CrossDomain}  & $23.77_{(0.42)}$  & $72.72_{(0.70)}$  & $38.87_{(0.52)}$  & $67.34_{(0.76)}$         & \multicolumn{1}{|c|}{$-$}          &      \multicolumn{1}{|c|}{$-$}                         &     \multicolumn{1}{|c|}{$-$}                        &    \multicolumn{1}{|c}{$-$}                         \\
   RelationNet+FWT$^\dagger$\cite{Tseng2020CrossDomain} & $23.95_{(0.20)}$          & $75.78_{(0.40)}$   & $38.68_{(0.30)}$           & $69.13_{(0.40)}$     & $65.55_{(0.40)}$         & $44.29_{(0.30)}$   & $40.18_{(0.40)}$         & $59.77_{(0.40)}$ \\
   GNN+FWT$^\dagger$\cite{Tseng2020CrossDomain}& $24.28_{(0.20)}$ & $87.07_{(0.40)}$  & $40.87_{(0.40)}$  & $78.02_{(0.40)}$  & $70.70_{(0.50)}$ & $49.66_{(0.40)}$ & $46.19_{(0.40)}$  & $64.97_{(0.50)}$       \\
  TPN+FWT $^\dagger$\cite{Tseng2020CrossDomain} & $21.22_{(0.10)}$ & $70.06_{(0.70)}$  & $36.96_{(0.40)}$  & $65.69_{(0.50)}$ & $66.75_{(0.50)}$  & $43.20_{(0.50)}$ & $34.03_{(0.40)}$  & $58.18_{(0.50)}$\\
     ATA$^\dagger$\cite{advTaskAug}  & $24.43_{(0.20)}$  & {${90.59}_{(0.30)}$}   & {${45.83}_{(0.30)}$}   & {${{83.75}}_{(0.40)}$} & $\underline{75.48}_{(0.40)}$ & {${55.08}_{(0.40)}$}   & {${49.14}_{(0.40)}$} & {${66.22}_{(0.50)}$} \\
HVM\cite{du2022hierarchical}  & $\mathbf{27.15}_{(0.45)}$ & $87.65_{(0.35)}$  & $42.05_{(0.34)}$           & $74.88_{(0.45)}$       & \multicolumn{1}{|c|}{$-$}          &      \multicolumn{1}{|c|}{$-$}                         &     \multicolumn{1}{|c|}{$-$}                        &    \multicolumn{1}{|c}{$-$}                         \\
ConFeSS \cite{das2022confess}  & $\underline{27.09}$ & $88.88$  & $48.85$           & $84.65$       & \multicolumn{1}{|c|}{$-$}          &      \multicolumn{1}{|c|}{$-$}                         &     \multicolumn{1}{|c|}{$-$}                        &    \multicolumn{1}{|c}{$-$}                         \\
  RDC-FT\cite{li2022ranking}      & ${25.48}_{(0.20)}$& $\underline{93.55}_{(0.30)}$ & $\underline{49.06}_{(0.30)}$ & $\underline{84.67}_{(0.30)}$    & $74.65_{(0.40)}$           &     $\underline{60.63}_{(0.40)}$                         &     $\underline{53.75}_{(0.50)}$                         &    $\underline{67.77}_{(0.40)}$                          \\
 AWCoL  & $24.50_{(0.33)}$ &	$\mathbf{99.59}_{(0.10)}$ &	$\mathbf{58.75}_{(0.60)}$&	$\mathbf{96.76}_{(0.27)}$ &	$\mathbf{92.56}_{(0.36)}$&	$\mathbf{67.31}_{(0.64)}$&	$\mathbf{62.94}_{(0.75)}$&	$\mathbf{86.23}_{(0.52)}$\\  
\hline

\end{tabular}} 
\label{table:5shot}
\vskip -.2in
\end{table*}

\paragraph{Datasets}
We conducted comprehensive experiments to evaluate the performance of AWCoL on eight CDFSL benchmark datasets. 
Following the standard CDFSL setting, we used Mini-ImageNet \cite{vinyals2016matching} as the source domain dataset in all experiments and employed the following eight datasets separately as the target domain datasets: ChestX \cite{wang2017chestx}, CropDiseases\cite{mohanty2016using}, ISIC \cite{tschandl2018ham10000}, EuroSAT \cite{helber2019eurosat}, Places \cite{zhou2017places}, Planatae \cite{van2018inaturalist}, Cars \cite{krause20133d} and CUB \cite{wah2011caltech}. We used the same train/validation/test splits as in \cite{guo2020broader} to ensure fair evaluation and comparison.

\paragraph{Implementation Details}
We use ResNet10 \cite{he2016deep} 
as the backbone network for the two prototypical classification models in AWCoL. 
The distance function employed in AWCoL is the squared L2 norm. 
We train each model in the source domain for 400 iterations with 100 randomly sampled FSL tasks 
and 15 query instances per class, using the Adam optimizer with a learning rate of 1e-3. 
We evaluate AWCoL on 600 randomly selected few-shot tasks in each target domain. 
For each task, we randomly sample 5 classes and 
randomly select 15 images per class as the query set. 
We fine-tune the models 
for a total of 100 iterations for each novel target task using the Adam optimizer with a learning rate of 1e-3. 
The trade-off parameter $\lambda$ is set to 1e-2, 
while $\alpha_{min}$, $\alpha_0$ (the initial value for $\alpha_m$), $\gamma$, and $\beta$ 
take the values of $0.1$, $0.5$, $0.99$ and $5$ respectively. 
In the first $\beta$ iterations of the fine-tuning procedure, 
we initialize the WMA prediction vectors of the fixed model with the current predictions of the updated model.

\subsection{Comparison Results}

\subsubsection{Learning with Few Shots}

We evaluate the proposed AWCoL method on the standard cross-domain 5-way 5-shot tasks.
We compare AWCoL with a set of representative FSL baselines (MatchingNet \cite{vinyals2016matching}, MAML \cite{finn2017model}, ProtoNet \cite{snell2017prototypical}, RelationNet \cite{sung2018learning}, MetaOpt \cite{lee2019meta}, GNN \cite{garcia2018fewshot} and TPN \cite{liu2018learning}) 
as well as 5 state-of-the-art CDFSL methods (FWT \cite{Tseng2020CrossDomain}, ATA \cite{advTaskAug}, 
HVM \cite{du2022hierarchical}, ConFeSS \cite{das2022confess} and RDC-FT \cite{li2022ranking}). 
FWT has been applied jointly with five standard FSL methods: MatchingNet, ProtoNet, RelationNet, GNN and TPN.
The comparison results are reported in Table~\ref{table:5shot}, where the top section of the table presents the results of the standard FSL methods and the bottom section presents the results of the CDFSL methods. 

The table shows that most CDFSL methods 
naturally outperform the FSL methods on all the datasets given their ability to handle the significant domain shift between the source and target domains. 
The proposed method AWCoL outperforms all the FSL and CDFSL methods 
on all the datasets except ChestX. 
On ChestX, AWCoL is slightly 
outperformed by a few other CDFSL methods. 
Nevertheless, the performance gains produced by AWCoL 
are notable exceeding 12\%, 17\% and 19\% on EuroSAT, Places and CUB datasets respectively. 
This highlights the 
superior performance of 
our proposed method over the existing state-of-the-art CDFSL methods in the literature.

\begin{table*}[t]
\caption{Mean classification accuracy (95\% confidence interval in brackets) on 4 target domain datasets for cross-domain 5-way 20-shot and 50-shot classification. ``$^\ast$'' indicates results reported in \cite{guo2020broader}.} 
\setlength{\tabcolsep}{2pt}	
\resizebox{\textwidth}{!}{
\begin{tabular}{l|l|l|l|l|l|l|l|l  }
\hline	
&   \multicolumn{2}{c|}{ChestX}         &   \multicolumn{2}{c|}{CropDiseases}   &           \multicolumn{2}{c|}{ISIC}             &     \multicolumn{2}{c}{EuroSAT}                  \\
  &  20-shot & 50-shot &  20-shot & 50-shot &  20-shot & 50-shot &  20-shot & 50-shot \\   
   
   \hline 
 MatchingNet$^\ast$\cite{vinyals2016matching} & $23.61_{(0.86)}$  &    $22.12_{(0.88)}$   & $76.38_{(0.67)}$ &    $58.53_{(0.73)}$ & $45.72_{(0.53)}$ &    $54.58_{(0.65)}$ & $77.10_{(0.57)}$ &    $54.44_{(0.67)}$          \\
   MAML$^\ast$\cite{finn2017model} & $27.53_{(0.43)}$    &    $-$   & $89.75_{(0.42)}$    &    $-$   & $52.36_{(0.57)}$    &   $-$   & $81.95_{(0.55)}$   &    $-$    \\
 ProtoNet$^\ast$\cite{snell2017prototypical}  & $28.21_{(1.15)}$   &    $29.32_{(1.12)}$   & $88.15_{(0.51)}$   &    $90.81_{(0.43)}$   & $49.50_{(0.55)}$        &    $51.99_{(0.52)}$   & $82.27_{(0.57)}$  &    $80.48_{(0.57)}$       \\
   MetaOpt$^\ast$\cite{lee2019meta}  & $25.53_{(1.02)}$   &    $29.35_{(0.99)}$   & $82.89_{(0.54)}$    &    $91.76_{(0.38)}$   & $49.42_{(0.60)}$   &    $54.80_{(0.54)}$   & $79.19_{(0.62)}$   &    $83.62_{(0.58)}$            \\

 RelationNet$^\ast$\cite{sung2018learning}     & $26.63_{(0.92)}$     &    $28.45_{(1.20)}$   & $80.45_{(0.64)}$  &    $85.08_{(0.53)}$   & $41.77_{(0.49)}$  &    $49.32_{(0.51)}$   & $74.43_{(0.66)}$  &    $74.91_{(0.58)}$     \\
\hline
 MatchingNet+FWT$^\ast$\cite{Tseng2020CrossDomain}  & $23.23_{(0.37)}$        &    $23.01_{(0.34)}$   & $74.90_{(0.71)}$         &    $75.68_{(0.78)}$   & $32.01_{(0.48)}$     &    $33.17_{(0.43)}$   & $63.38_{(0.69)}$  &    $62.75_{(0.76)}$     \\
 ProtoNet+FWT$^\ast$\cite{Tseng2020CrossDomain}   & $26.87_{(0.43)}$   &    $30.12_{(0.46)}$   & $85.82_{(0.51)}$   &    $87.17_{(0.50)}$   & $43.78_{(0.47)}$   &    $49.84_{(0.51)}$   & $75.74_{(0.70)}$   &    $78.64_{(0.57)}$        \\
 RelationNet+FWT$^\ast$\cite{Tseng2020CrossDomain} & $26.75_{(0.41)}$    &    $27.56_{(0.40)}$   & $78.43_{(0.59)}$    &    $81.14_{(0.56)}$   & $43.31_{(0.51)}$  &    $46.38_{(0.53)}$   & $69.40_{(0.64)}$   &    $73.84_{(0.60)}$  \\
    HVM\cite{du2022hierarchical}  & ${ {30.54}}_{(0.47)}$  &    ${32.76}_{(0.46)}$   & ${95.13}_{(0.35)}$   &    $\underline{97.83}_{(0.33)}$   & ${54.97}_{(0.35)}$            &    ${61.71}_{(0.32)}$   & ${84.81}_{(0.34)}$      &    ${87.16}_{(0.35)}$    \\
	ConFeSS \cite{das2022confess}  & $\underline{33.57}$ & $\mathbf{39.02}$  & $\underline{95.34}$           & ${97.56}$      & $\underline{60.10}$ & $\underline{65.34}$  & $\underline{90.40}$           & $\underline{92.66}$               \\
  AWCoL  & $\mathbf{34.13}_{(0.47)}$  &    $\underline{36.72}_{(0.48)}$   & $\mathbf{99.93}_{(0.03)}$  &    $\mathbf{99.98}_{(0.02)}$   & $\mathbf{72.40}_{(0.52)}$  &    $\mathbf{73.40}_{(0.51)}$   & $\mathbf{98.98}_{(0.12)}$  &    $\mathbf{98.73}_{(0.16)}$   \\  
\hline
\end{tabular}} 
\label{table:20shot}
\vskip -.2in
\end{table*}

\subsubsection{Learning with Higher Shots}

We also conducted experiments to evaluate the performance of the proposed method on higher-shot tasks in the target domain.
In particular, we evaluate 
AWCoL on 5-way 20-shot and 5-way 50-shot learning tasks on four target-domain datasets: 
ChestX, CropDiseases, ISIC and EuroSAT. The results are presented in Table~\ref{table:20shot}, 
where the top section of the table reports results of 
the standard FSL methods and the bottom section 
reports results of the CDFSL methods.

Similar to the comparisons on 5-shot tasks, 
all the CDFSL methods (except FWT) outperform 
the in-domain FSL methods across all datasets 
for both 20-shot and 50-shot tasks. 
It is also clear from the table that AWCoL outperforms all the other FSL and CDFSL methods on 
the CropDiseases, ISIC and EuroSAT datasets
for both 20-shot and 50-shot learning tasks. 
The performance gains are notable exceeding 12\% and 8\% on ISIC dataset in the cases of 20-shot and 50-shot learning tasks respectively. AWCoL is slightly outperformed on ChestX by ConFess in the case of 50-shot tasks. Nevertheless our proposed method still obtains the second best result and best result on ChestX for the cases of 50-shot and 20-shot tasks respectively.

\subsection{Ablation Study}

\begin{table*}[t]
\centering
\caption{
	Ablation study results in terms of mean classification accuracy (95\% confidence interval within brackets) for cross-domain 5-way 5-shot classification tasks.}
\setlength{\tabcolsep}{3pt}	
\resizebox{\textwidth}{!}{
\begin{tabular}{l|l|l|l|l|l|l|l|l}
\hline	
 & ChestX                    & CropDisea.                & ISIC                       & EuroSAT                  & Places                   & Planatae                    & Cars                     & CUB                      \\
   \hline 
AWCoL                  &  $24.50_{(0.33)}$ &	$\mathbf{99.59}_{(0.10)}$ &	$\mathbf{58.75}_{(0.60)}$&	${96.76}_{(0.27)}$ &	${92.56}_{(0.36)}$&	$\mathbf{67.31}_{(0.64)}$&	$\mathbf{62.94}_{(0.75)}$&	$\mathbf{86.23}_{(0.52)}$\\ \hline

$\; - \text{w/o}$ Co-Learn.  & $ 24.08_{ (0.42)}$&$ 81.65_{( 0.62 )}$ & $42.93_{(0.60)}$ & $72.72_{(0.72)}$ & $69.15_{(0.73)}$& $46.16_{(0.65)}$& $42.60_{(0.72)}$&	$56.35_{(0.73)}$ \\
$\; - \text{w/o }$ Alt. Update  &$\textbf{24.79}_{ (0.40) }$&$ 84.31_{(0.63)}$&$44.42_{ (0.61) }$&$ 75.34_{ (0.69)}$                          &$68.91_{(0.70)}$&	$49.40_{(0.67)}$	&$44.53_{(0.77)}$ &	$58.63_{(0.72)}$       \\
$\; - \text{w/o }$ WMA                  &$ 24.47_{ (0.44)}$&$ 84.19_{ (0.62)}$&$ 43.96_{(0.63)}$&$ 75.31_{ (0.70)}$  & $68.85_{(0.71)}$&	$48.19_{(0.67)}$	&$40.05_{(0.68)}$ &$58.05_{(0.70)}$                                \\
$\; - \text{w/o } \mathcal{L}_m^{co}$          &           $23.45_{ (0.41)}$ & $ 84.74_{(0.62)}$&$ 41.83_{(0.62)}$ & $71.16_{(0.69)}$ & $69.77_{(0.73
)}$&$47.72_{(0.69)}$&	$43.30_{(0.75)}$&	$56.07_{(0.71)}$\\                                      

$\; - \text{w/o } \mathcal{L}_m^{\mathcal{N}}$    & $23.22_{(0.33)}$                                 & $99.50_{(0.12)}$&$ 57.82_{ (0.65)}$ & $\mathbf{96.86}_{ (0.24)}$           &      	 $\mathbf{93.13}_{(0.33)}$&	${66.90}_{(0.65)}$&	${62.90}_{(0.69)}$&	${85.28}_{(0.51)}$\\  
                                  
$\; - \text{w/o }$ Adapt. Weight               &$ 23.62_{(0.34)}$                                  &$ 98.52_{(0.09)}$&$ 57.45_{(0.62 )}$& $96.65_{(0.27)}$                    &	${\bf 93.06}_{(0.36)}$&	${62.23}_{(0.75)}$&	${61.54}_{(0.77)}$&	${84.61}_{(0.52)}$\\

$\; - \text{with}\, \mathcal{L}_m^{S}$                             & $22.28_{(0.26)}$          & $99.22_{(0.12)}$          &$54.98_{(0.63)}$          &$ 93.97_{(0.36)}$     & $86.94_{(0.49)}$ &	$46.38_{(0.64)}$&	$37.28_{(0.70)}$&	$77.77_{(0.57)} $    \\ \hline

\end{tabular}}
\label{table:ablation}
\vskip -.20in
\end{table*}

In order to investigate the contribution of each component of the proposed method, 
we conducted an ablation study to compare the proposed AWCoL with its seven variants: 
(1) ``$- \text{w/o }$ Co-Learn.", which drops the adaptive co-learning component and fine-tunes each model independently using its own WMA predictions. 
(2) ``$- \text{w/o }$ Alt. Update", which updates the two models 
simultaneously
instead of using the alternating updates strategy. 
(3) ``$- \text{w/o }$ WMA", which drops the WMA prediction generating strategy 
by setting $\alpha_m=1$.
(4) ``$- \text{w/o } \mathcal{L}_m^{co}$", which drops the adaptive co-learning loss term ($\mathcal{L}_m^{co}$) from the total loss function. 
(5) ``$- \text{w/o } \mathcal{L}_m^{\mathcal{N}}$", which drops the negative pseudo-labeling regularization loss term ($\mathcal{L}_m^{\mathcal{N}}$) from the total loss function. 
(6) ``$- \text{w/o } $ Adapt. Weight", which drops the adaptive weighting component of AWCoL by setting all $w_{[x]}=1$.
(7) ``$- \text{with } \mathcal{L}_m^{S}$", which add a cross-entropy loss term on the labeled support instances to the total loss function for each model. 

We compare the proposed AWCoL
with all of its seven variants using the cross-domain 5-way 5-shot learning tasks 
on all the eight datasets and report the results in Table~\ref{table:ablation}. 
From the table, we can see that dropping any component from the proposed full model
results in performance degradation in almost all the 
48 cases except for only 4 cases: the ``$- \text{w/o }$ Alt. Update" variant on ChestX,
the ``$- \text{w/o }$ Adapt. Weight" variant on Places, 
and the ``$- \text{w/o } \mathcal{L}_m^{\mathcal{N}}$" variant on EuroSAT and Places. 
Even with the exception on ChestX, 
the ``$- \text{w/o }$ Alt. Update" variant demonstrates substantial performance drops
from the full AWCoL method on all the other datasets, ranging from 14\% (on ISIC) to 27\% (on CUB).  
Such consistent performance drops across many datasets
validate the essential contribution of each corresponding component of the proposed AWCoL. 
Meanwhile, with an additional loss term on the support instances, 
the ``$- \text{with } \mathcal{L}_m^{S}$" variant 
produces much inferior results to AWCoL across all the eight datasets. 
The possible reason is that the labeled support instances have already been used
to produce the class prototypes for the cross-entropy loss on the query instances in AWCoL, 
hence an additional cross-entropy loss $\mathcal{L}_m^{S}$ may 
overfit the limited support instances.


\section{Conclusion}

In this paper, we proposed 
a weighted adaptive co-learning method to address 
the challenging cross-domain few-shot learning problem. 
The proposed method fine-tunes two prototypical classification models independently pre-trained in the source 
domain for the target FSL task.
In each co-learning iteration, a weighted moving average strategy is deployed to 
generate probability predictions for the query instances using each model separately. 
The predictions of the two models are then combined to produce 
positive pseudo-labels, negative pseudo-labels, and adaptive weights for the query instances. 
We adopted an alternating fine-tuning mechanism to 
update each model separately 
by minimizing the weighted cross-entropy loss
over the pseudo-labeled query instances while maximizing a similar cross-entropy loss 
with negative pseudo-labels to penalize false predictions. 
We conducted extensive experiments on eight CDFSL benchmark datasets.
The results demonstrated the effectiveness of the proposed simple method 
comparing with the state-of-the-art baselines.

{\small
\bibliographystyle{splncs04}
\bibliography{egbib}

\begin{thebibliography}{10}
\providecommand{\url}[1]{\texttt{#1}}
\providecommand{\urlprefix}{URL }
\providecommand{\doi}[1]{https://doi.org/#1}

\bibitem{adler2020cross}
Adler, T., Brandstetter, J., Widrich, M., Mayr, A., Kreil, D., Kopp, M.,
  Klambauer, G., Hochreiter, S.: Cross-domain few-shot learning by
  representation fusion. In: arXiv preprint arXiv:2010.06498 (2020)

\bibitem{chen2020diversity}
Chen, M., Fang, Y., Wang, X., Luo, H., Geng, Y., Zhang, X., Huang, C., Liu, W.,
  Wang, B.: Diversity transfer network for few-shot learning. In: AAAI
  Conference on Artificial Intelligence (2020)

\bibitem{das2022confess}
Das, D., Yun, S., Porikli, F.: Confe{SS}: A framework for single source
  cross-domain few-shot learning. In: International Conference on Learning
  Representations (ICLR) (2022)

\bibitem{du2022hierarchical}
Du, Y., Zhen, X., Shao, L., Snoek, C.G.M.: Hierarchical variational memory for
  few-shot learning across domains. In: International Conference on Learning
  Representations (ICLR) (2022)

\bibitem{finn2017model}
Finn, C., Abbeel, P., Levine, S.: Model-agnostic meta-learning for fast
  adaptation of deep networks. In: International Conference on Machine Learning
  (ICML) (2017)

\bibitem{guo2020broader}
Guo, Y., Codella, N.C., Karlinsky, L., Codella, J.V., Smith, J.R., Saenko, K.,
  Rosing, T., Feris, R.: A broader study of cross-domain few-shot learning. In:
  European Conference on Computer Vision (ECCV) (2020)

\bibitem{guo2019spottune}
Guo, Y., Shi, H., Kumar, A., Grauman, K., Rosing, T., Feris, R.: Spottune:
  transfer learning through adaptive fine-tuning. In: Conference on Computer
  Vision and Pattern Recognition (CVPR) (2019)

\bibitem{he2016deep}
He, K., Zhang, X., Ren, S., Sun, J.: Deep residual learning for image
  recognition. In: Proceedings of the IEEE conference on computer vision and
  pattern recognition (2016)

\bibitem{helber2019eurosat}
Helber, P., Bischke, B., Dengel, A., Borth, D.: Eurosat: A novel dataset and
  deep learning benchmark for land use and land cover classification. In: IEEE
  Journal of Selected Topics in Applied Earth Observations and Remote Sensing
  (2019)

\bibitem{islam2021dynamic}
Islam, A., Chen, C.F.R., Panda, R., Karlinsky, L., Feris, R., Radke, R.J.:
  Dynamic distillation network for cross-domain few-shot recognition with
  unlabeled data. In: Advances in Neural Information Processing Systems
  (NeurIPS) (2021)

\bibitem{jeong2020ood}
Jeong, T., Kim, H.: Ood-maml: Meta-learning for few-shot out-of-distribution
  detection and classification. In: Advances in Neural Information Processing
  Systems (NeurIPS) (2020)

\bibitem{krause20133d}
Krause, J., Stark, M., Deng, J., Fei-Fei, L.: 3d object representations for
  fine-grained categorization. In: International Conference on Computer Vision
  workshops (2013)

\bibitem{lee2019meta}
Lee, K., Maji, S., Ravichandran, A., Soatto, S.: Meta-learning with
  differentiable convex optimization. In: Conference on Computer Vision and
  Pattern Recognition (CVPR) (2019)

\bibitem{li2022ranking}
Li, P., Gong, S., Wang, C., Fu, Y.: Ranking distance calibration for
  cross-domain few-shot learning. In: Conference on Computer Vision and Pattern
  Recognition (CVPR) (2022)

\bibitem{lim2019fast}
Lim, S., Kim, I., Kim, T., Kim, C., Kim, S.: Fast autoaugment. In: Advances in
  Neural Information Processing Systems (NeurIPS) (2019)

\bibitem{liu2018learning}
Liu, Y., Lee, J., Park, M., Kim, S., Yang, E., Hwang, S., Yang, Y.: Learning to
  propagate labels: Transductive propagation network for few-shot learning. In:
  International Conference on Learning Representations (ICLR) (2019)

\bibitem{mohanty2016using}
Mohanty, S.P., Hughes, D.P., Salath{\'e}, M.: Using deep learning for
  image-based plant disease detection. In: Frontiers in plant science (2016)

\bibitem{garcia2018fewshot}
Satorras, V.G., Estrach, J.B.: Few-shot learning with graph neural networks.
  In: International Conference on Learning Representations (ICLR) (2018)

\bibitem{snell2017prototypical}
Snell, J., Swersky, K., Zemel, R.: Prototypical networks for few-shot learning.
  In: Advances in neural information processing systems (NIPS) (2017)

\bibitem{sung2018learning}
Sung, F., Yang, Y., Zhang, L., Xiang, T., Torr, P.H., Hospedales, T.M.:
  Learning to compare: Relation network for few-shot learning. In: Conference
  on Computer Vision and Pattern Recognition (CVPR) (2018)

\bibitem{tschandl2018ham10000}
Tschandl, P., Rosendahl, C., Kittler, H.: The ham10000 dataset, a large
  collection of multi-source dermatoscopic images of common pigmented skin
  lesions. In: Scientific data (2018)

\bibitem{Tseng2020CrossDomain}
Tseng, H.Y., Lee, H.Y., Huang, J.B., Yang, M.H.: Cross-domain few-shot
  classification via learned feature-wise transformation. In: International
  Conference on Learning Representations (ICLR) (2020)

\bibitem{van2018inaturalist}
Van~Horn, G., Mac~Aodha, O., Song, Y., Cui, Y., Sun, C., Shepard, A., Adam, H.,
  Perona, P., Belongie, S.: The inaturalist species classification and
  detection dataset. In: Conference on Computer Vision and Pattern Recognition
  (CVPR) (2018)

\bibitem{vinyals2016matching}
Vinyals, O., Blundell, C., Lillicrap, T., Wierstra, D., et~al.: Matching
  networks for one shot learning. In: Advances in neural information processing
  systems (NIPS) (2016)

\bibitem{wah2011caltech}
Wah, C., Branson, S., Welinder, P., Perona, P., Belongie, S.: The caltech-ucsd
  birds-200-2011 dataset. California Institute of Technology (2011)

\bibitem{advTaskAug}
Wang, H., Deng, Z.H.: Cross-domain few-shot classification via adversarial task
  augmentation. In: International Joint Conference on Artificial Intelligence
  (IJCAI) (2021)

\bibitem{wang2017chestx}
Wang, X., Peng, Y., Lu, L., Lu, Z., Bagheri, M., Summers, R.M.: Chestx-ray8:
  Hospital-scale chest x-ray database and benchmarks on weakly-supervised
  classification and localization of common thorax diseases. In: Conference on
  Computer Vision and Pattern Recognition (CVPR) (2017)

\bibitem{xu2022co}
Xu, R., Xing, L., Shao, S., Liu, B., Zhang, K., Liu, W.: Co-learning for
  few-shot learning. In: Neural Processing Letters (2022)

\bibitem{yin2021hierarchical}
Yin, C., Wu, K., Che, Z., Jiang, B., Xu, Z., Tang, J.: Hierarchical graph
  attention network for few-shot visual-semantic learning. In: International
  Conference on Computer Vision (ICCV) (2021)

\bibitem{zhang2018metagan}
Zhang, R., Che, T., Ghahramani, Z., Bengio, Y., Song, Y.: Metagan: An
  adversarial approach to few-shot learning. In: Advances in Neural information
  processing systems (NeurIPS) (2018)

\bibitem{zhou2017places}
Zhou, B., Lapedriza, A., Khosla, A., Oliva, A., Torralba, A.: Places: A 10
  million image database for scene recognition. In: IEEE Transactions on
  Pattern Analysis and Machine Intelligence (PAMI) (2017)

\end{thebibliography}
}

\end{document}